\documentclass[letterpaper, 10 pt, conference]{ieeeconf}  % Comment this line out if you need a4paper

\IEEEoverridecommandlockouts                              % This command is only needed if 
\usepackage{amsmath}
\usepackage{amsfonts}
\usepackage{booktabs}
\usepackage{graphicx}
\usepackage{tabularx}

\usepackage{pgfplots}
\pgfplotsset{compat=1.18}
\usepackage{caption}
\usepackage{titletoc}
\usepackage{multirow}

\usepackage{listings}
\usepackage{xcolor}
\usepackage{svg}

\newcommand{\rev}[1]{\textcolor{black}{#1}}

\title{\LARGE \bf
SLoSa: Small, Low-Power, and Safe Foundation Models for Robotic Task Planning
}

\author{Weizhe Xu$^{1}$, Jialiang Fan$^{1}$, Essien Taylor$^{2}$, Zijun Wang$^{1}$, Russell Joseph$^{2}$, and Fanxin Kong$^{1}$\\
\textit{$^{1}$University of Notre Dame \quad $^{2}$Northwestern University}\\
\{wxu3, jfan5, zwang58, fkong\}@nd.edu\\
etaylor@u.northwestern.edu, rjoseph@northwestern.edu
}

\begin{document}

\maketitle
\thispagestyle{empty}
\pagestyle{empty}

\begin{abstract}
Safe robotic task planning requires planners that can reliably achieve task goals while satisfying safety constraints.
Classical planners search over explicit symbolic task representations and can
provide formal guarantees with respect to encoded safety constraints, but require rigid input formats.
Their computational cost also increases rapidly with the search space.
In contrast, large language models (LLMs) offer flexible task interfaces and strong generalization capabilities, but they are large, power-hungry, and pre-trained with little robot-specific safety data and knowledge.
Thus, they can be difficult to deploy locally on robotic systems with limited resources.
To address these limitations, we present SLoSa, a neuro-symbolic compression and fine-tuning framework for training foundation-model-based robotic task planners with \underline{S}mall size, \underline{Lo}w power demand, and \underline{Sa}fety awareness. 
The framework first constructs a safety-aware dataset using classical planners, an advanced teacher LLM, and formal verifiers, followed by two training stages: supervised fine-tuning and reinforcement learning (RL) fine-tuning guided by a verification-based reward machine to enforce safety alignment.
\rev{We evaluate SLoSa across four Planning Domain Definition Language (PDDL) domains, two household simulation environments, local edge platforms, and a desktop robotic arm. 
Across the PDDL tasks, verifier-guided post-training raises the success rate of compact 4B–14B planners to 70.5–76.0\%, while INT4 deployment on a Jetson Orin requires 17.85–26.41 W of average power. 
Results in household simulators and on the physical robot further demonstrate the applicability of SLoSa across diverse robotic task settings.}
\end{abstract}

\section{Introduction}

Robotic task planning converts a high-level goal into an executable sequence of actions~\cite{guo2023recent}.
In safety-critical settings, reaching the goal is not enough: every intermediate state must also satisfy requirements such as ordering, mutual exclusion, and resource constraints~\cite{jha2023neuro,xu2024assuring}.
A plan that eventually places an object correctly can still be unacceptable if it enters a restricted region, manipulates a fragile object in the wrong order, or violates a required inspection step.
We therefore study \emph{safety-constrained task planning}: generating plans that are executable and goal-reaching while satisfying explicit trajectory-level constraints.

Classical planners offer a principled solution when a correct symbolic \rev{task specification} is available.
Systems such as Fast Downward and OPTIC search over states defined by action preconditions and effects~\cite{helmert2006fast,benton2012temporal}; PDDL3 expresses trajectory constraints, and tools such as VAL check a candidate plan against the \rev{formal specification}~\cite{gerevini2005plan,howey2004val}.
This formal structure provides an auditable safety criterion, but it comes with two practical costs.
First, tasks in complicated environments can induce exponentially growing search spaces as the number of objects and planning horizon increase, making symbolic search difficult to scale~\cite{kwon2025fast}.
Second, users must provide rigid symbolic inputs and carefully engineered domain \rev{specifications}.
Imitation and reinforcement learning can amortize planning into compact policies~\cite{choudhury2018data,mandlekar2023human,acharya2023neurosymbolic,jiang2019task}.
However, learned policies are commonly tied to particular task distributions, and their outputs do not inherit a formal guarantee without an explicit \rev{specification} and verifier.

LLM planners offer a complementary strength: they can interpret high-level
task descriptions and directly generate candidate action
sequences~\cite{jansen2020visually,brohan2023rt}.
Recent systems ground language into symbolic specifications or integrate LLMs
with planners and formal verifiers~\cite{quartey2025verifiably,rabiei2025ltlcodegen,wu2025selp}.
An inference-time verifier can reject an invalid plan
and prompt the model to revise it~\cite{zhou2024isr,yang2024plug}.
However, this process leaves the base planner unchanged, so repeated failures
increase token usage, latency, and computation.
Moreover, general-purpose LLMs can be too large for memory- and power-limited
robot computers.
These limitations motivate a planner that combines safety-aware plan
generation, compact local inference, and flexible task interfaces.

We introduce \textbf{SLoSa}, a neuro-symbolic post-training framework for this combination.
Rather than keeping the verifier only in the inference loop, SLoSa uses formal feedback to change the compact planner itself.
First, SLoSa constructs behavior-changing safety-aware training data by retaining planning instances in which the supplied constraint rejects the baseline plan while a verified alternative remains feasible.
Second, compression-aware supervised fine-tuning (SFT) adapts compact, quantized LLMs using the verified plans.
Third, verifier-guided Group Relative Policy Optimization (GRPO) refines the planner with a reward hierarchy that separates unparseable outputs, precondition violations, safety violations, goal failures, and successful plans.
At inference, SLoSa combines local plan generation with verifier feedback to enforce the supplied safety constraints while reducing the tokens spent on iterative repair.

\rev{We evaluate SLoSa using three compact open-source LLM backbones across four PDDL domains that cover manipulation, transportation, object transfer, and tool use, together with separately trained planners for safety-adapted VirtualHome and ALFWorld to evaluate the applicability of the same post-training pipeline beyond PDDL-based planning.}
The results show consistent improvements from safety-aware post-training
across model sizes, task domains, and input representations.
We further evaluate local inference on an NVIDIA RTX 4080 and a Jetson Orin,
and demonstrate the complete vision-grounded planning pipeline on a desktop
\rev{6-degree-of-freedom (6-DoF)} robotic arm.
Our contributions are summarized as follows: 
\begin{itemize}
    \item We introduce SLoSa, a neuro-symbolic post-training framework that
    improves compact LLM planners through behavior-changing safety-aware data
    construction, compression-aware supervised fine-tuning, and
    verifier-guided GRPO.
    \item \rev{We construct safety-constrained evaluation tasks spanning four PDDL planning domains and two adapted household simulation benchmarks, with formally verified plans and task-specific trajectory constraints.}
    \item We conduct extensive experiments across multiple model sizes, task
    domains, and input representations, showing consistent improvements from
    safety-aware post-training. We further validate local edge deployment,
    verifier-guided inference, and physical robotic-arm execution.
\end{itemize}

\section{Related Work}

\textbf{Classical, learned, and neuro-symbolic task planning.}
Classical planners search over explicit \rev{task representations} and can validate a plan against its goals and constraints~\cite{helmert2006fast,benton2012temporal,gerevini2005plan,howey2004val}.
Their explicit semantics make failures interpretable, but formal specification construction is labor-intensive, and tasks in complicated environments can induce exponentially growing search spaces as the number of objects and the planning horizon increase~\cite{kwon2025fast}.
Imitation learning and reinforcement learning instead learn policies or search guidance from demonstrations and interaction~\cite{choudhury2018data,mandlekar2023human,acharya2023neurosymbolic,jiang2019task}.
But they are generally optimized for a specific training distribution.
\rev{SLoSa adopts this general direction but uses symbolic planners and validators primarily as sources of training supervision, while a compact neural model generates the initial plan at deployment.}

\textbf{LLM planning with formal grounding and runtime assurance.}
LLMs have been used to generate actions, select skills, and translate natural-language requests into planning specifications~\cite{jansen2020visually,brohan2023rt}.
LIMP grounds open-ended robot instructions into symbolic representations from which verifiable behaviors can be synthesized~\cite{quartey2025verifiably}, while 
\cite{rabiei2025ltlcodegen, wu2025selp} translate language into syntactically valid temporal logic for downstream motion planning.
Other systems retain a verifier in the inference loop: ISR-LLM translates natural-language tasks into PDDL and iteratively refines long-horizon plans, while the Safety Chip enforces LTL constraints around an LLM-driven agent~\cite{zhou2024isr,yang2024plug}.
These approaches provide important assurance at the interface between language and execution, but a weak base planner can still require repeated generation and verification.
SLoSa instead uses formal validation during post-training to improve the first-pass plan quality of the compact planner itself.
At deployment, the resulting planner can still be paired with an ISR-style verifier and repair loop, retaining formal guarantees for validator-accepted plans while reducing regeneration tokens.

\textbf{Safety post-training and compact deployment.}
Quantization and parameter-efficient fine-tuning reduce the memory and adaptation cost of LLMs~\cite{zhu2024survey,egashira2024exploiting,chavan2023one,li2024loftq}, but compression alone does not optimize satisfaction of task-level safety constraints.
SafeGen-LLM uses verified PDDL supervision to improve safety generalization across properties, domains, and input formats~\cite{fan2026safegen}.
SLoSa studies a \rev{complementary} question: whether verifier knowledge can be transferred into small, low-power LLM planners that improve safe plan generation with lower memory use and less dependence on repeated repair.

\section{Preliminaries}
\label{sec:preliminaries}

\textbf{Safety-constrained robotic task planning.}
We consider a high-level robotic planning problem
$P=(S,A,s_0,G,C)$,
where $S$ is the symbolic state space, $A$ is the action set, $s_0$ is the
initial state, $G$ is the goal, and $C$ is the set of safety constraints.
Each action $a\in A$ has preconditions $\mathrm{Pre}(a)$ and effects
$\mathrm{Eff}(a)$, and represents a high-level robot skill such as pick,
move, or place.

A plan is a finite action sequence $\tau=[a_1,\ldots,a_T]$.
A plan is executable, denoted by $\mathrm{Exec}(\tau)$, if
$s_{i-1}\models\mathrm{Pre}(a_i)$ for all $i=1,\ldots,T$.
If $\mathrm{Exec}(\tau)$ holds, applying the action effects induces a complete
trajectory
$\rho(\tau)=(s_0,s_1,\ldots,s_T)$.
For an executable plan, goal achievement and safety are defined as
$
\mathrm{Goal}(\tau)\iff s_T\models G,
\mathrm{Safe}(\tau)\iff \rho(\tau)\models C.$

\textbf{PDDL3 constraints and plan verification.}
We represent symbolic planning problems using PDDL and encode hard safety
requirements as PDDL3 trajectory constraints.
A PDDL3 domain defines predicates and actions, whereas a problem specifies
objects, the initial state, the goal, and trajectory constraints.

Given a generated output, the verifier applies its checks in a fixed order,
and the first failed check determines the outcome.
An output that cannot be parsed or contains undefined actions or objects is a
\emph{format error}.
A parsed plan that violates an action precondition is a
\emph{precondition violation}.
Only after obtaining a complete executable trajectory does the verifier
evaluate $C$; an executable plan that violates $C$ is a
\emph{safety violation}.
A safe and executable plan is a \emph{goal failure} if it does not reach $G$,
and a \emph{success} otherwise.
This ordered procedure assigns exactly one category to every output and is
used for training data filtering, reward design, and evaluation.

\textbf{SLoSa objective.}
SLoSa stands for \underline{S}mall size, \underline{Lo}w power demand, and \underline{Sa}fety awareness. 
Here, safety awareness means compliance with the formal task specification defined above, while low-power operation is an empirical deployment objective rather than an assumed property.
The goal of SLoSa robotic task planning is to train a compact LLM planner that generates safety-compliant plans with low deployment and inference cost.

\section{SLoSa}

\begin{figure*}[t]
    \centering
    \includegraphics[width=0.85\textwidth]{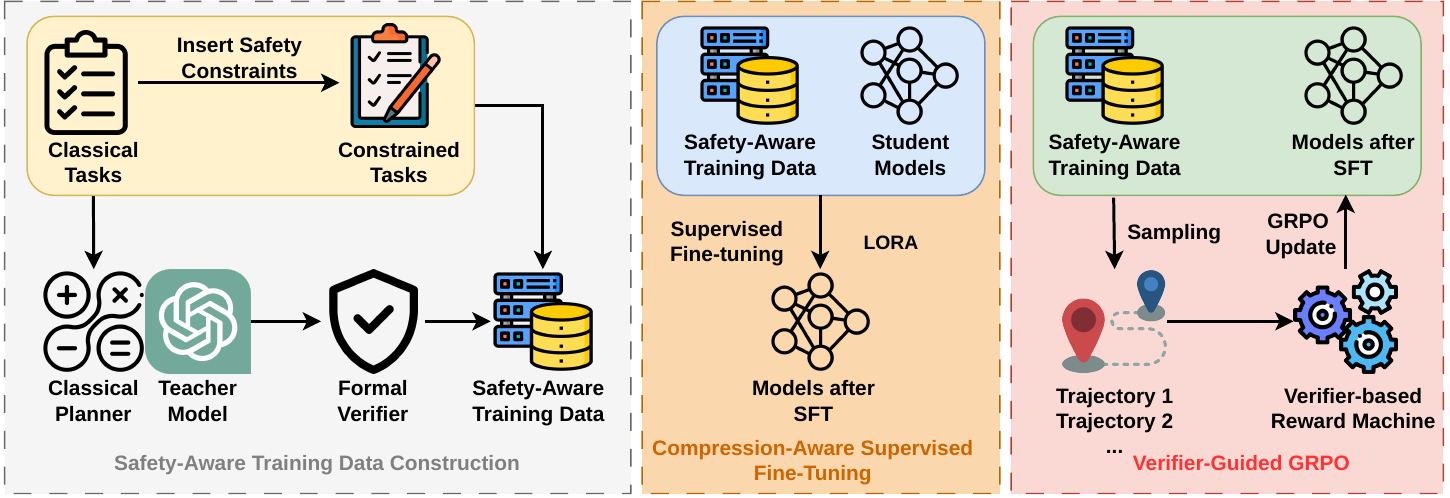}
    % \includesvg[width=0.85\textwidth]{figures/SLoCe_Robot_Overview_5}
    \vspace{-0.5em}
    \caption{
    Overview of SLoSa. 
    Classical planners, teacher models, and a formal verifier construct safety-aware data. 
    Student LLMs are trained by SFT and then refined by GRPO with verifier-based rewards.
    }
    \vspace{-1.5em}
    \label{fig:slocerobot_overview}
\end{figure*}

SLoSa is a neuro-symbolic compression and fine-tuning framework for safety-constrained robotic task planning. 
It trains compact, quantized LLM planners using formally verified supervision and reward signals.
As shown in Figure~\ref{fig:slocerobot_overview}, SLoSa has three stages. 
First, it constructs safety-aware training data using classical planners, teacher models, and formal validators. 
Only plans that pass formal validation are used as positive training targets.
Second, it applies SFT to compact and quantized LLMs using verified plans.
The model takes PDDL, structured text, or natural language as input and outputs a grounded action sequence. 
Third, it applies verifier-guided GRPO, where formal verification gives error categories and classical planning gives progress signals for dense reward design. 
Together, these stages use symbolic planning and verification to guide neural model compression and fine-tuning, so the final planner is easier to deploy and more likely to generate safety-compliant plans.

\subsection{Safety-Aware Training Data Construction}

The first stage of SLoSa constructs behavior-changing, safety-constrained
training instances for SFT and GRPO.
We use four robotics-inspired PDDL domains: Blocksworld, Ferry, Grippers, and
Spanner, covering manipulation, transportation, object transfer, and tool use.

For each base problem $P_0$, a classical planner generates a baseline plan
$\tau_0$, which is verified as executable and goal reaching.
We then introduce a domain-grounded PDDL3 constraint $c$ to obtain a
constrained problem $P_c$.
The constraint is derived from the domain semantics and the actions and objects
appearing in the task, following one of the four patterns summarized in
Table~\ref{tab:safety_patterns}.
For example, in Spanner, an ordering constraint requiring \texttt{nut1}
to be tightened before \texttt{nut2} is encoded as
\texttt{(sometime-before (tightened nut2) (tightened nut1))}.
A plan that tightens both nuts in the reverse order may still reach the
task goal, but violates the inserted trajectory constraint.

\begin{table}[t]
    \centering
    \caption{Safety patterns used to construct behavior-changing constraints.}
    \vspace{-0.5 em}
    \label{tab:safety_patterns}
    \footnotesize
    \setlength{\tabcolsep}{3pt}
    \renewcommand{\arraystretch}{0.95}
    \begin{tabular}{p{0.24\linewidth} p{0.29\linewidth} p{0.34\linewidth}}
        \toprule
        \textbf{Pattern} & \textbf{PDDL3 construct} &
        \textbf{Robotic interpretation} \\
        \midrule
        Ordering safety &
        \texttt{sometime-before} &
        A required event must precede another event. \\

        Staging safety &
        \texttt{sometime}, \texttt{sometime-before} &
        The robot must pass through a safe intermediate state. \\

        Resource restriction &
        \texttt{always}, \texttt{not} &
        Unsafe, reserved, or faulty resources cannot be used. \\

        Mandatory inspection &
        \texttt{sometime} &
        The robot must visit a checkpoint or complete an inspection. \\
        \bottomrule
    \end{tabular}
    \renewcommand{\arraystretch}{1.0}
    \vspace{-1.5em}
\end{table}

A planner then solves $P_c$ to obtain an alternative plan $\tau_c$.
Let $V(P,\tau)$ denote the verifier outcome for plan $\tau$ under problem $P$.
We retain an instance only if \rev{
\(V(P_0,\tau_0)=V(P_c,\tau_c)=\mathrm{success}\)
and
\(V(P_c,\tau_0)=\mathrm{safety\ violation}\).}
Thus, the added constraint invalidates the baseline plan while preserving a
verified alternative, ensuring that the constraint changes the required
planning behavior rather than acting as a decorative condition.
We therefore refer to the retained instances as \emph{behavior-changing}.
This definition is relative to the selected baseline plan $\tau_0$ and does
not imply that the formal specification captures every real-world hazard.

\rev{We assign each constrained problem a difficulty level solely according to the length of its verified plan, denoted by $d(P_c)=\operatorname{bin}(|\tau_c|)$.
The Easy, Medium, and Hard levels therefore differ only in planning-horizon length.
}
This measure represents the effective planning horizon under the safety
constraint rather than difficulty inferred from problem-size parameters.

Each constrained problem $P_c$ is converted into task prompts using PDDL or template-generated natural language.
The prompt is paired with the verified constrained plan $\tau_c$ as a positive
SFT example and is used without a target plan during GRPO.
\rev{For 20\% of the constrained problems, a teacher model (GPT-5.4-mini) generates additional
candidate plans to increase demonstration diversity.
These plans are retained only if they receive a success label from the same
formal verifier.}

Across the dataset, we use three input formats: PDDL, structured natural language, and narrative natural language. 
Each constrained problem contributes one PDDL prompt and one natural-language prompt rendered in either structured or narrative form.
These representations preserve the same underlying symbolic problem while exposing the planner to different task interfaces.

\subsection{Compression-Aware Supervised Fine-Tuning}

The safety-aware dataset provides verified plans, but the pretrained compact models do not know the required planning format, action semantics, or safety-constrained behavior. 
We therefore use SFT as the first training stage. 
The goal is to give the compact planner a strong initialization before GRPO, so that it can generate parseable and executable plans that are close to the verified constrained plans.

Given a positive training set
$\mathcal{D}_{+}=\{(x_i,\tau_i)\}_{i=1}^{N}$,
where $x_i$ is a task prompt constructed from a constrained planning problem
and $\tau_i$ is its verifier-accepted plan, we minimize
{\small
\begin{equation*}
\mathcal{L}_{\text{SFT}}(\theta)
=
-\frac{1}{N}
\sum_{i=1}^{N}
\log p_{\theta}(\tau_i \mid x_i).
\end{equation*}
}
The target final output is always a grounded action sequence, with one action per line. 
SFT uses only verifier-accepted plans as positive targets.
To support the SLoSa objective, we fine-tune compact LLMs with quantization and parameter-efficient adapters. 
The base model is loaded in low precision, and only the adapter parameters are trained.

\subsection{Verifier-Guided GRPO}

SFT learns from verified target plans but does not directly learn from the
model's own failures.
We therefore refine the SFT model using GRPO, with the formal verifier serving
as a programmatic reward function.
This avoids training a separate reward model or critic.

For each prompt group $g$, the current policy samples $K$ candidate plans
$\{\tau_{g,1},\ldots,\tau_{g,K}\}$.
The verifier assigns each candidate one of the five mutually exclusive
outcomes defined in Section~\ref{sec:preliminaries} and computes a scalar
reward $r_{g,i}=r(x_g,\tau_{g,i})$.

\textbf{Verifier-guided reward.}
Format errors receive a fixed reward of $-1$, while successful plans receive
$1$.
For precondition violations, safety violations, and goal failures, we define
negative reward intervals $I_c=[\ell_c,h_c]$ satisfying
\[
\begin{aligned}
-1 < \ell_{\mathrm{pre}} \leq h_{\mathrm{pre}}
< \ell_{\mathrm{safe}} \leq h_{\mathrm{safe}}
< \ell_{\mathrm{goal}} \leq h_{\mathrm{goal}} < 0.
\end{aligned}
\]
We order the reward intervals according to the verification pipeline and failure severity.
A precondition violation cannot produce a complete execution trace, while, among executable plans, a safety violation is treated as more severe than a safe plan that fails only to reach the goal.

Planner-based progress distinguishes candidates within the same failure
category.
For each evaluation state, we construct a residual planning problem by replacing the original initial state with the evaluation state while retaining the same action model and goal.
A classical planner configured for shortest-plan search then solves the residual problem.
For a precondition violation, the evaluation state is the state before the failed action.
For a safety violation, it is the first state at which the violation is detected, or the terminal state for constraints evaluated over the complete trajectory.
For a goal failure, it is the terminal state.
We define $d_{\mathrm{init}}$ as the length of the shortest plan from the
original initial state to the goal and $d_{\mathrm{res}}$ as the length of the
shortest plan from the evaluation state to the goal:
\[
\phi_{\mathrm{dist}}
=
\frac{d_{\mathrm{init}}-d_{\mathrm{res}}}
     {d_{\mathrm{init}}+d_{\mathrm{res}}+\epsilon}.
\]
A larger $\phi_{\mathrm{dist}}$ indicates greater progress toward the goal.
If the classical planner cannot solve the residual problem within the time
limit, we assign the lowest progress value.

For a failure in category
$c\in\{\mathrm{pre},\mathrm{safe},\mathrm{goal}\}$, the reward is
\[
r(x,\tau)
=
\ell_c+(h_c-\ell_c)
\frac{\phi_{\mathrm{dist}}+1}{2}.
\]
The verifier determines the reward interval, while the classical-planner
distance orders candidates only within that interval.
We omit a plan-length term from successful plans because the objective is to
increase validator acceptance rather than optimize plan cost.

\textbf{Active-group GRPO updates.}
For each prompt group $g$, GRPO computes the normalized advantages
{\small
\begin{equation*}
A_i
=
\frac{
r(x,\tau_i)-\operatorname{mean}_{j} r(x,\tau_j)
}{
\operatorname{std}_{j} r(x,\tau_j)+\epsilon
}.
\end{equation*}
}
If all rewards are identical, the group provides no relative preference signal
and is marked inactive.
Inactive groups are excluded from both the policy and KL terms.
If a nominal step contains no active group, its optimizer and learning-rate
scheduler updates are skipped, while the step counter and data cursor continue
to advance.
No additional candidates are sampled, preserving the data order and generation
budget.

\textbf{Difficulty curriculum.}
GRPO uses the difficulty levels defined during data construction from the
lengths of the verified constrained plans.
Early training emphasizes problems with shorter plans, after which the
proportion of medium and hard problems gradually increases.
This curriculum allows the policy to first learn relative preferences over
shorter planning horizons before training on longer ones.

\section{Experiments}
\label{sec:experiments}
\rev{To evaluate the effectiveness and versatility of SLoSa across diverse robotic planning settings, we consider PDDL planning domains, household simulation benchmarks, and a real-world robotic arm.}
We further measure local deployment cost and conduct ablation studies of the
safety-aware post-training pipeline.

\subsection{Experimental Setup}

\textbf{PDDL tasks.}
\rev{We evaluate SLoSa on four robotics-inspired domains: Blocksworld, Ferry, Grippers, and Spanner.
For each domain, we generate 250 constrained planning problems, each represented by one PDDL prompt and one natural-language prompt rendered in either structured or narrative form, yielding 500 prompt instances per domain.
Across all domains, the SFT and GRPO sets each contain 1,000 prompts derived from disjoint sets of 500 underlying planning problems.
Validation and test splits are also disjoint at the underlying problem level to avoid cross-format leakage.}

\textbf{Models, training, and decoding.}
We study Qwen3-4B\footnote{\texttt{unsloth/Qwen3-4B-unsloth-bnb-4bit}},
Llama-8B\footnote{\texttt{unsloth/Meta-Llama-3.1-8B-Instruct-bnb-4bit}},
and Qwen3-14B\footnote{\texttt{unsloth/Qwen3-14B-unsloth-bnb-4bit}}.
All models are loaded with 4-bit quantization and adapted using LoRA with rank
32, scaling factor 64, and dropout 0.05.
SFT uses a learning rate of $2\times10^{-4}$ and an effective batch size of 8.
GRPO uses fused AdamW with a learning rate of $1\times10^{-5}$, $\beta_1=0.9$, $\beta_2=0.999$, $\epsilon=10^{-8}$, no weight decay, and a maximum gradient norm of 1.0.
The learning rate decays linearly over 500 effective optimizer updates without warmup.
Each nominal step contains four prompt groups, with $K=4$ candidate plans sampled per prompt at temperature 0.9 and a KL coefficient of 0.02.
GRPO follows a three-stage easy-to-hard curriculum with Easy/Medium/Hard sampling ratios of $6/3/1$, $1/6/3$, and $1/3/6$, over 125, 250, and 125 steps, respectively.
Checkpoints are saved every 125 steps, and the checkpoint with the highest validation success is selected for evaluation.
For verifier-guided rewards, we use
$I_{\mathrm{pre}}=[-0.9,-0.7]$,
$I_{\mathrm{safe}}=[-0.6,-0.4]$, and
$I_{\mathrm{goal}}=[-0.3,-0.1]$;
format errors and successful plans receive fixed rewards of $-1$ and $1$, respectively.

At evaluation, all models use greedy decoding without intermediate
chain-of-thought output.
Qwen3 thinking mode is disabled during both training and evaluation.
The PT, SFT, and GRPO models share the same test split and output format.

\textbf{Household benchmarks.}
To test whether SLoSa applies beyond PDDL, we additionally use fully observed,
non-interactive adaptations of VirtualHome~\cite{puig2018virtualhome} and
ALFWorld~\cite{shridhar2020alfworld}. For each benchmark, we construct a
separate safety-aware training and test set.
We match the dataset size and difficulty distribution to the PDDL setup.
We then independently train the Qwen3-4B SFT and GRPO checkpoints from the same pretrained initialization.
Each model must emit a complete plan without intermediate environment feedback
or replanning. ALFWorld-FO-Safe exposes the complete symbolic state and is not
the standard partially observed, interactive ALFWorld protocol.

\textbf{Metrics and reporting.}
We use the verifier-defined success rate as the primary metric and report the
mutually exclusive failure outcomes for diagnosis.
For deployment, end-to-end latency includes generation and parsing, whereas throughput measures decoding only.

\subsection{Planning Performance}

Table~\ref{tab:training_results} reports the results on the current 200-instance PDDL evaluation. 
\rev{PT denotes the original pretrained compact backbone before SLoSa post-training. Because these models have not yet been adapted to the task-specific action vocabulary and strict output format, their failures are dominated by formatting or grounding errors. In contrast, the larger directly prompted LLMs more often produce parseable outputs, although many of their plans still fail the verifier.}
SFT produces the largest change from PT: success rises from 0\% to 59.0--67\%.
GRPO then improves success to 70.5\%, 74.0\%, and 76.0\% for the 4B, 8B, and 14B models, respectively. 
These consistent gains across model families and scales demonstrate that SLoSa
improves the safety-constrained planning capability of compact LLMs.
Directly prompted GPT-5.6 Sol and Claude Sonnet 5 achieve 26.5\% and 31.0\% success, respectively, showing that model scale alone does not replace safety-aware post-training on these formally constrained tasks.

\begin{table*}[t]
    \centering
    \caption{PDDL planning outcomes across post-training stages. Values are percentages over the current 200-instance test set; outcome columns are mutually exclusive. GPT and Claude models are evaluated by direct prompting, whereas the compact models are shown across their available post-training stages.
    }
    \label{tab:training_results}
    % \small
    \footnotesize
    \renewcommand{\arraystretch}{0.88}
    \setlength{\tabcolsep}{8pt}
    \begin{tabular}{llccccc}
        \toprule
        Model & Stage &  Success $\uparrow$ & Goal fail $\downarrow$ &
        Safety violation $\downarrow$ & Precondition violation $\downarrow$ &
        Format error $\downarrow$ \\
        \midrule
        GPT-5.6 Sol & Direct & 26.5 & 0.0 & 0.5 & 73.0 & 0.0 \\
        \midrule
        Claude Sonnet 5 & Direct & 31.0 & 3.5 & 0.0 & 55.5 & 10.0 \\
        \midrule
          \multirow{3}{*}{Qwen3-4B}
          & PT   & 0.0 & 1.5 & 0.0 & 0.0 & 98.5 \\
          & SFT  & 59.0 & 1.5 & 0.5 & 36.5 & 2.5 \\
          & GRPO & 70.5 & 1.5 & 0.5 & 25.0 & 2.5 \\
          \midrule
          \multirow{3}{*}{Llama-8B}
          & PT   & 0.0 & 5.0 & 0.0 & 70.0 & 25.0 \\
          & SFT  & 62.0 & 5.5 & 1.0 & 30.5 & 1.0 \\
          & GRPO & 74.0 & 7.5 & 0.0 & 18.0 & 0.5 \\
          \midrule
          \multirow{3}{*}{Qwen3-14B}
          & PT   & 0.0 & 0.5 & 0.0 & 0.0 & 99.5 \\
          & SFT  & 67.0 & 2.5 & 0.0 & 30.5 & 0.0 \\
          & GRPO & \textbf{76.0} & 1.5 & 0.0 & 22.0 & 0.5 \\
        \bottomrule
    \end{tabular}
    \vspace{-1.5em}
\end{table*}

\begin{figure}[t]
    \centering
    % \includesvg[width=0.8\columnwidth]
    % {figures/qwen3_4b_format_and_stacked_outcomes_single_column_top_labels}
    \includegraphics[width=0.8\columnwidth]
    {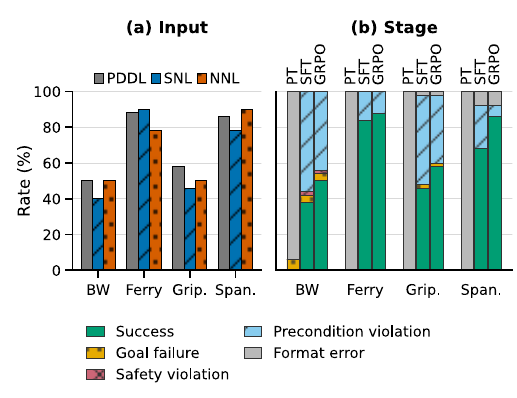}
    \vspace{-1 em}
    \caption{Domain and input-format analysis of Qwen3-4B.
\rev{Panel (a) compares the GRPO-refined model across PDDL, structured natural language (SNL), and narrative natural language (NNL) inputs.
Panel (b) shows the distribution of outcomes for PT, SFT, and GRPO in each domain.}
    }
    \label{fig:domain_input_analysis}
    \vspace{-1.5em}
  \end{figure}

\rev{Figure~\ref{fig:domain_input_analysis} provides a domain- and input-level breakdown of the Qwen3-4B results. 
The outcome distributions show that the benefits of post-training are consistent across the four PDDL domains, although Blocksworld remains the most challenging. 
The GRPO-refined model also achieves comparable performance across PDDL, structured natural language, and narrative natural language inputs, indicating robustness to different task representations.}

\subsection{Contribution of Safety-Aware Post-Training}
We isolate the contribution of safety-aware SFT data using Qwen3-4B.
All variants use equally sized positive-only datasets and receive no
contrastive negative examples.
U-SFT uses unconstrained positive plans.
OC-SFT uses constraint-compliant plans from a balanced mixture: in 50\% of
the problems, the constraint does not alter the nominal solution, while the
remaining 50\% are behavior-changing.
BC-SFT uses only behavior-changing instances in which the constraint
invalidates the nominal plan while preserving a verified alternative.
All variants are trained using SFT alone, allowing the comparison to isolate
the effect of safety-aware data composition without confounding from GRPO.

We distinguish task completion from safety compliance using three metrics.
\emph{Success} requires a well-formed, executable, goal-reaching, and
constraint-compliant plan.
\emph{Goal reach} counts well-formed and executable plans that reach the
goal, regardless of constraint compliance.
\emph{Conditional safety violation} is the fraction of goal-reach outputs
that violate at least one supplied safety constraint.

\begin{table}[t]
    \centering
    \caption{Positive-only safety-aware data ablation on Qwen3-4B.
    Brackets report bootstrap 95\% confidence intervals for success.}
    \label{tab:data_ablation}
    % \scriptsize
    % \setlength{\tabcolsep}{2.8pt}
    \begin{tabular}{lccc}
        \toprule
        Data &
        \shortstack{Success\\(95\% CI) $\uparrow$} &
        \shortstack{Goal\\reach $\uparrow$} &
        \shortstack{Cond. safety\\violation $\downarrow$} \\
        \midrule
        U-SFT  & 3.5\% [1.7, 7.0]    & 17.5\% & 80.0\% \\
        OC-SFT & 53.0\% [46.1, 59.8] & 56.0\% & 5.4\%  \\
        BC-SFT & \textbf{59.0\% [52.1, 65.6]} & \textbf{59.5\%} & \textbf{0.8\%}  \\
        \bottomrule
    \end{tabular}
    \vspace{-1.5em}
\end{table}

Constraint-aware supervision substantially improves success over U-SFT.
Using behavior-changing instances further reduces unsafe goal-reaching plans,
showing that the proposed data filtering teaches the model to change its
planning behavior in response to safety constraints rather than merely imitate
goal-reaching solutions.

% \subsection{Applicability Beyond PDDL}
\subsection{Applicability Across Planning Benchmarks}

We apply the same post-training pipeline separately to VirtualHome-Safe and ALFWorld-FO-Safe. 
Each row in Table~\ref{tab:household_applicability} uses a Qwen3-4B planner trained on benchmark-specific data; the comparison therefore tests whether SLoSa remains effective beyond PDDL domains.

\begin{table}[t]
    \centering
    \caption{Success (\%) of separately trained Qwen3-4B planners on two
    household benchmarks. SFT and GRPO use benchmark-specific training data;
    each checkpoint generates one complete plan without intermediate feedback.}
    \label{tab:household_applicability}
    \small
    \setlength{\tabcolsep}{3.0pt}
    \begin{tabular}{lccc}
        \toprule
        Benchmark & PT & SFT & GRPO \\
        \midrule
        VirtualHome-Safe & 0.0\% & 74.5\% & 80.5\% \\
        ALFWorld-FO-Safe & 0.0\% & 76.0\% & 85.0\% \\
        \bottomrule
    \end{tabular}
    \vspace{-2 em}
\end{table}

Across both benchmarks, SFT substantially improves planning success over the pre-trained model, while verifier-guided GRPO provides a further gain.
These consistent trends show that the proposed post-training pipeline remains effective across planning benchmarks with different task representations and execution environments.

\begin{figure*}[t]
    \centering
    \begin{minipage}[t]{0.62\textwidth}
        \centering
        \includegraphics[width=\linewidth]{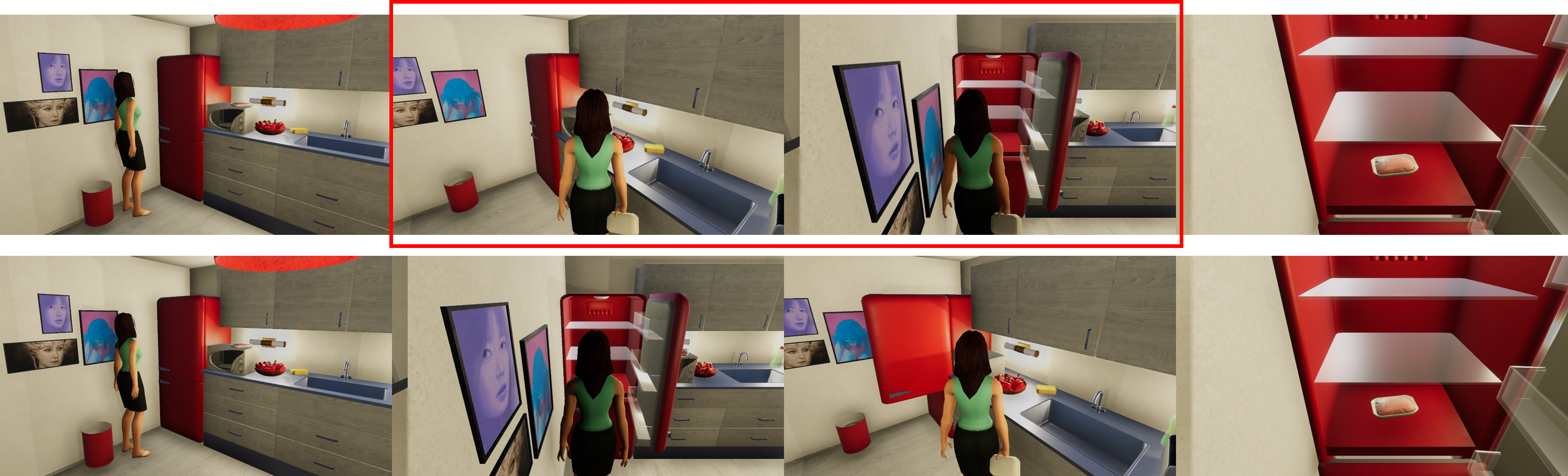}
        % \includesvg[width=\linewidth]{figures/virtualhome_2}
        \vspace{-1.5em}
        \caption*{(a) VirtualHome-Safe}
    \end{minipage}
    \hfill
    \begin{minipage}[t]{0.36\textwidth}
        \centering
        \includegraphics[width=\linewidth]{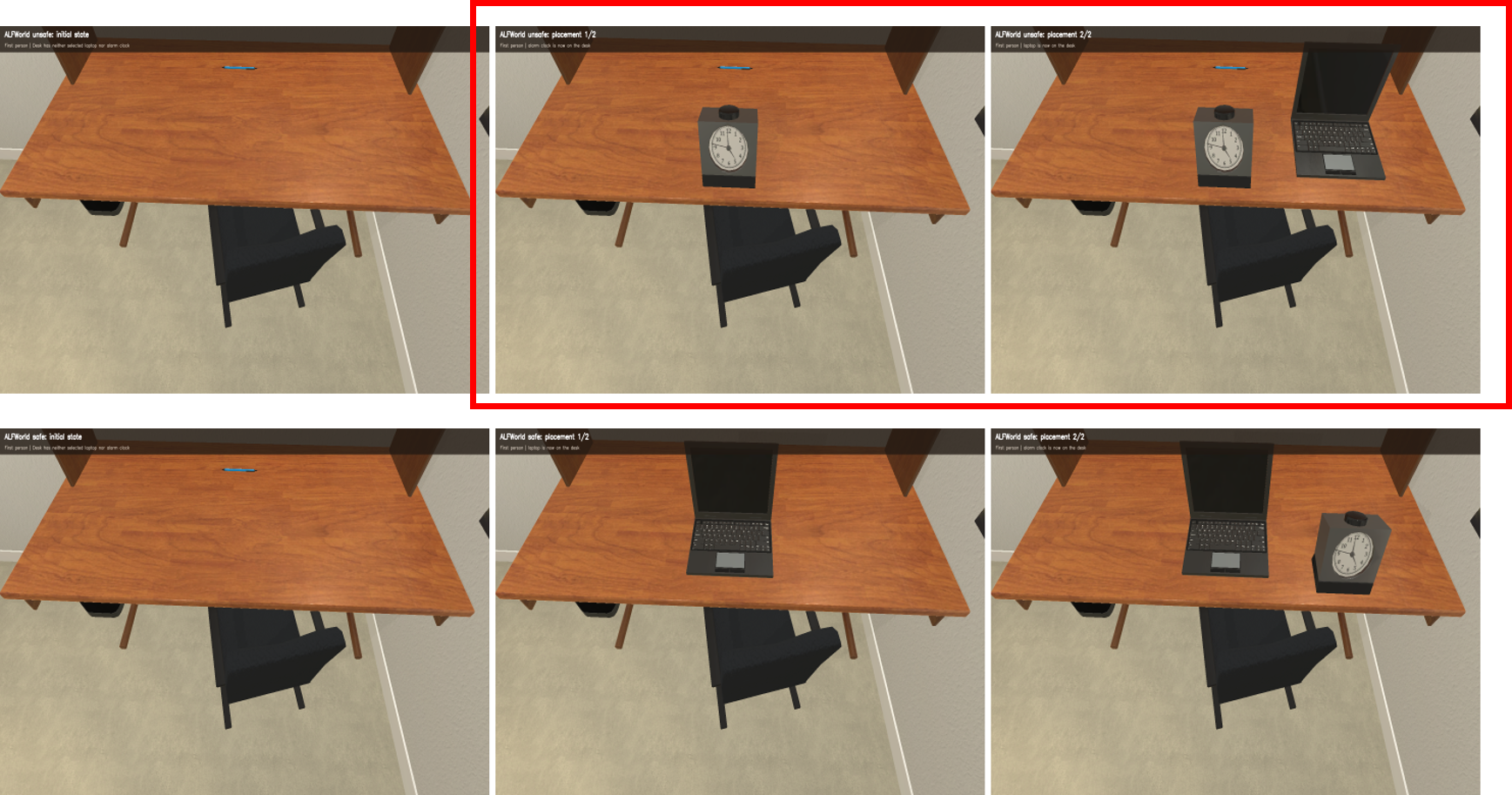}
        % \includesvg[width=\linewidth]{figures/alfworld_2}
        \vspace{-1.5em}
        \caption*{(b) ALFWorld-FO-Safe}
    \end{minipage}
    \vspace{-0.5em}
    \caption{Representative safety-constrained planning problems in two embodied
simulators.
In each panel, the upper sequence shows the original baseline plan, which
completes the task but violates the added safety constraint, whereas the lower
sequence shows a safety-satisfying plan for the same task.
The red box highlights the critical steps that cause the constraint violation.
In panel (a), the task is to place the salmon in the refrigerator, with the
refrigerator required to be opened before the salmon is picked up.
In panel (b), the task is to place the clock and laptop on the table, with the
laptop required to be placed before the clock.
The examples assess high-level symbolic safety compliance rather than physical collision safety.}
    \label{fig:simulator_safety_sequences}
    \vspace{-1 em}
\end{figure*}

Figure~\ref{fig:simulator_safety_sequences} presents representative
safety-constrained problems constructed in VirtualHome-Safe and
ALFWorld-FO-Safe.
For each problem, the upper row shows the original baseline plan, which
completes the task but violates the added safety constraint, whereas the lower
row shows a safety-satisfying plan for the same task.
The examples illustrate how the introduced constraints change the required
planning behavior in both simulation environments.

\subsection{Edge Deployment and Repair Cost}

We evaluate INT4 local inference on an NVIDIA RTX 4080 (16GB) and a NVIDIA Jetson AGX Orin Developer Kit (64GB)
using batch size one and a single concurrent request.
Each model is evaluated on the same 100 PDDL prompts over three runs, following
20 warmup queries.
Table~\ref{tab:edge_efficiency} shows that all three compact planners fit and
run locally on both devices.
On the Jetson Orin, the models require 3.89--11.50\,GB of peak memory and
achieve 3.55--5.01 tokens/s while drawing 17.85--26.41\,W on average.
Llama-8B provides the lowest latency and highest throughput, whereas Qwen3-4B
has the lowest memory, power, and energy requirements.
These results demonstrate that SLoSa planners can be deployed entirely on
resource-constrained edge hardware without cloud inference.

\begin{table*}[t]
    \centering
    \caption{INT4 local-inference efficiency on an RTX 4080 and a Jetson Orin.
Latency includes generation and parsing, whereas throughput
measures generation only.
Energy per accepted plan divides the measured inference energy, including inter-request overhead, by the number of plans.
Model loading, warmup, and validation energy are excluded.
    }
    \label{tab:edge_efficiency}
    % \scriptsize
    % \setlength{\tabcolsep}{3.8pt}
    \begin{tabular}{llccccccc}
        \toprule
        Hardware & Model & P50 (s) & P90 (s) & Tok./s & Peak mem. (GB) &
        Avg. power (W) & Energy/ plan (J) \\
        \midrule
        \multirow{3}{*}{RTX 4080}
        & Qwen3-4B  & 6.65 & 8.66  & 11.73 & 3.80  & 71.16  & 464.75 \\
        & Llama-8B  & 6.01 & 7.77  & 13.19 & 5.91  & 88.14  & 516.70 \\
        & Qwen3-14B & 8.03 & 11.11 & 9.77  & 11.22 & 122.90 & 983.78 \\
        \midrule
        % \addlinespace[1pt]
        \multirow{3}{*}{Jetson Orin}
        & Qwen3-4B & 18.28 & 23.53 & 4.30 & 3.89 & 17.85 & 316.33  \\
        & Llama-8B & 15.88 & 20.01 & 5.01 & 6.06 & 23.14 & 355.18  \\
        & Qwen3-14B & 22.00 & 30.04 & 3.55 & 11.50 & 26.41 & 581.14  \\
        \bottomrule
    \end{tabular}
    \vspace{-1.5em}
\end{table*}

\begin{figure}[t]
    \centering
    \includegraphics[width=0.95\linewidth]{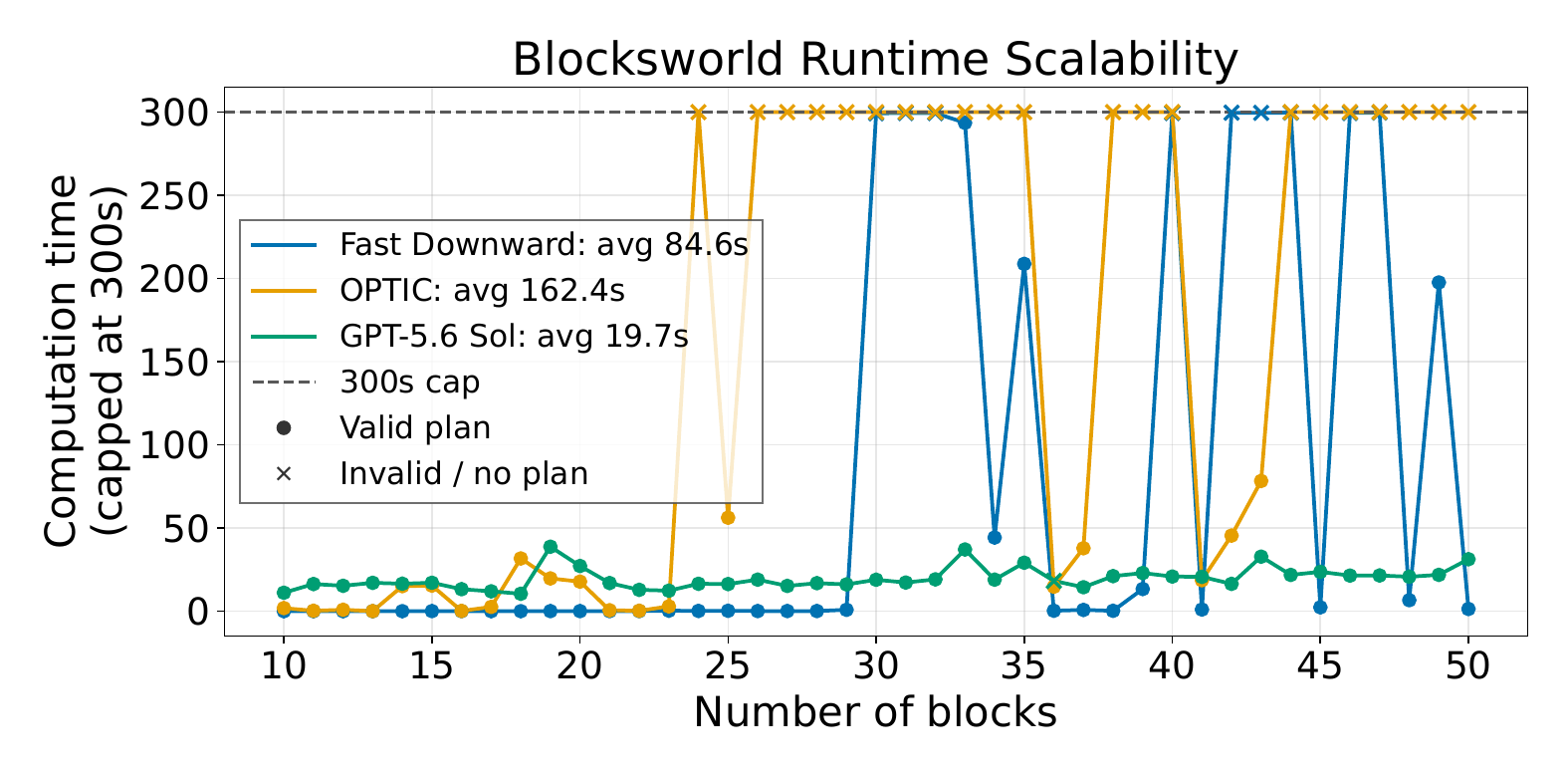}
    \vspace{-0.5 em}
    \caption{Runtime scalability on 41 Blocksworld instances with 10--50 blocks.
Fast Downward and OPTIC show highly variable search times, while GPT-5.6 Sol
maintains comparatively stable generation time.
Markers distinguish valid plans from invalid/no-plan outcomes, and the dashed
line denotes the 300-s timeout.
    }
    \label{fig:runtime_scalability}
    \vspace{-1 em}
\end{figure}

Figure~\ref{fig:runtime_scalability} compares runtime scaling on 41 Blocksworld instances with 10--50 blocks.
Fast Downward and OPTIC exhibit increasingly variable runtimes as the problem size grows, with many larger instances reaching the 300-s timeout.
Some larger instances are solved faster because search difficulty depends on the specific initial and goal configurations, rather than only on the number of blocks.
In contrast, GPT-5.6 Sol maintains comparatively stable generation time across the evaluated range.
This comparison motivates combining efficient LLM-based plan generation with
symbolic planning and formal verification, as pursued by SLoSa.

\rev{We then insert the three Llama-8B checkpoints into
SafePilot~\cite{xu2024assuring}, which uses verifier feedback to repair
rejected plans for at most three rounds.
SafePilot improves final success while ensuring that only plans passing all
verifier checks are accepted.
Table~\ref{tab:safepilot} further shows that SLoSa reduces the cost of this
verification-guided repair: stronger post-training requires fewer iterations
to obtain a verifier-accepted plan and consequently reduces token usage.}

\begin{table}[t]
    \centering
    \caption{\rev{Verification-guided inference with SafePilot over the 200-instance
    test set and at most three rounds. Improvement is measured relative to first-pass
    success before repair. The iteration count includes the initial generation.}}
    \label{tab:safepilot}
    \small
    \setlength{\tabcolsep}{4pt}
    \begin{tabular}{lcccc}
        \toprule
        Model & Final succ. $\uparrow$ & Gain $\uparrow$ &
        Avg. iter. $\downarrow$ & Avg. tokens $\downarrow$ \\
        \midrule
         PT
        & 0.5\% & +0.5 pp & 2.995 & 5185.81 \\
         SFT
        & 68.0\% & +6.0 pp & 1.735 & 2179.52 \\
         GRPO
        & \textbf{77.5\%} & +3.5 pp & \textbf{1.500} & \textbf{1782.87} \\
        \bottomrule
    \end{tabular}
    \vspace{-1em}
\end{table}

\subsection{Desktop Robot-Arm Demonstration}

The physical experiment implements Blocksworld using a desktop 6-DoF robotic
arm and a third-person camera.
Each task is specified by an image of the initial block configuration, an image
of the desired goal configuration, and a natural-language description of the
safety constraints.
A vision-language model grounds these inputs into a safety-constrained PDDL3
problem by extracting the objects, initial state, goal predicates, and safety constraints.
Qwen3-4B SFT or GRPO then generates a high-level plan from the resulting symbolic specification.

Before execution, we apply a SafePilot-style verifier-in-the-loop procedure.
The formal verifier checks each candidate plan against the problem.
If the plan is rejected, the verification feedback is returned to the planner
for revision; only a verifier-accepted plan is passed to the robotic-arm
controller, where symbolic actions are mapped to predefined manipulation
skills.
This workflow prevents invalid high-level action sequences from being sent
directly to the robot, thereby reducing the risk of unsafe motions and damage.

We evaluate each planner on 30 Blocksworld trials.
Table~\ref{tab:robot_results} reports verifier acceptance and physical
completion separately.
\rev{Verifier-}rejected plans are never executed, while low-level failures denote cases in which a verifier-accepted plan cannot be completed because limited execution precision causes block-stacking failure or the resulting stack exceeds the robot arm's reachable workspace.
This experiment demonstrates the practical feasibility of deploying SLoSa as part of a complete high-level planning and verification pipeline in a real-world robotic environment.

\begin{table}[t]
    \centering
    \caption{Blocksworld results on the desktop robotic arm.
    Each planner is evaluated on 30 planning trials; only verifier-accepted
    plans are executed on the robot.}
    \label{tab:robot_results}
    \small
    \setlength{\tabcolsep}{5pt}
    \begin{tabular}{lcc}
        \toprule
        Metric & SFT & Full GRPO \\
        \midrule
          Verifier-accepted plans (/30) & 17 & 19 \\
          \rev{Verifier-rejected plans} (/30) & 13 & 11 \\
          % Semantic violations (/30) & 13 & 11 \\
          Successful executions (/30) & 13 & 14 \\
          Low-level failures (/30) & 4 & 5 \\
        % \midrule
        % \multicolumn{2}{l}{Known-unsafe candidates rejected (/10)} & \\
        \bottomrule
    \end{tabular}
    \vspace{-1.5em}
\end{table}

\begin{figure}[t]
    \centering
    % \includesvg[width=0.9\columnwidth]{figures/robotic_arm_3}
    \includegraphics[width=0.9\columnwidth]{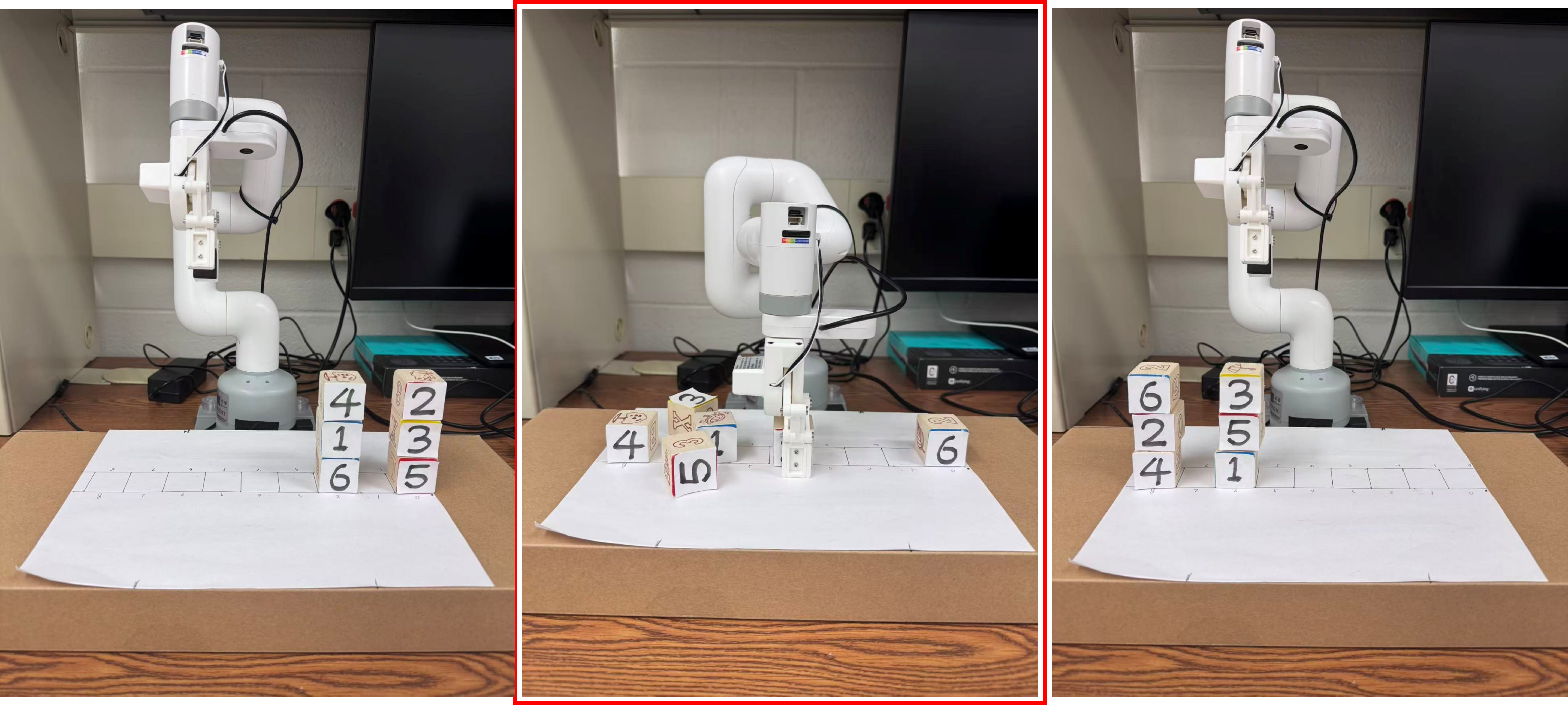}
    \caption{Representative Blocksworld execution with the desktop robotic arm.
    From left to right: the initial state, a failed physical outcome caused by
    block-placement error, and a successful goal state.
    % The middle and right panels show outcomes from separate executions of
    % verifier-accepted high-level plans.
    The physical failure results from limited manipulation precision rather
    than a violation of the encoded symbolic constraints.}
    \label{fig:robot_arm_safety_sequences}
    \vspace{-1.5em}
\end{figure}

\section{Discussion and Limitations}
SLoSa uses symbolic planning and formal verification as sources of training feedback rather than as restrictions on the deployment interface.
The resulting planners accept formal language, structured natural-language, and narrative natural-language inputs.
By transferring formal planning feedback into compact LLMs while preserving flexible task interfaces, SLoSa addresses the gap identified in the
Introduction: safety-aware planning, compact local inference, and flexible task
inputs within a single framework.

SLoSa focuses on high-level task planning.
Its safety supervision is derived from symbolic action models and formal
verifier outcomes, and the reported safety performance measures compliance
with the corresponding high-level constraints.
The evaluation does not cover perception errors, motion-level failures, or
unmodeled physical risks.
Long-horizon domains such as Blocksworld also remain challenging, and the
desktop-arm evaluation uses one workcell and fixed low-level skills.
Future work will study richer robotic tasks and tighter integration with
perception, motion planning, and runtime safety mechanisms.

\section{Conclusion}

We introduced SLoSa, a neuro-symbolic post-training framework for compact,
safety-constrained robotic task planning.
By using classical planners, a teacher LLM, and formal verifiers for training
supervision, SLoSa improves executable, goal-reaching, and safety-compliant
plan generation across model sizes, domains, and input formats.
Results on household benchmarks, SafePilot integration, and local deployment
further demonstrate applicability beyond PDDL and reduced inference overhead.
\rev{Overall, SLoSa provides a practical way to combine flexible LLM-based plan generation with symbolic planning and formal verification.}

% \clearpage

\bibliographystyle{IEEEtran}
\bibliography{example}

\end{document}